\documentclass{article}

\usepackage{times,uweetr}
\usepackage{graphicx}
\usepackage{paralist}
\usepackage{framed}
\usepackage{subfigure}
\usepackage[numbers]{natbib}
\usepackage{amsmath, amsthm, amssymb}
\usepackage{algorithm}
\usepackage{algorithmic}
\usepackage[dvipdfm]{hyperref}

\newtheorem{thm}{Theorem}
\newtheorem{lma}{Lemma}
\newtheorem{cor}{Corollary}
\newtheorem{prp}{Proposition}

\def\Naive{Na\"\i ve }
\def\Gain{\operatorname{Gain}}

\begin{document}

\title{Interactive Submodular Set Cover}
\author{Andrew Guillory \\
Computer Science and Engineering \\
University of Washington \\
\tt{guillory@cs.washington.edu} \\
\and Jeff Bilmes \\
Electrical Engineering \\
University of Washington \\
\tt{bilmes@ee.washington.edu}}
\reportmonth{May}
\reportyear{2010}
\reportnumber{0001}
\makecover
\maketitle

\begin{abstract}
  We introduce a natural generalization of submodular set cover and
  exact active learning with a finite hypothesis class (query
  learning).  We call this new problem interactive submodular set
  cover.  Applications include advertising in social networks with
  hidden information.  We give an approximation guarantee for a novel
  greedy algorithm and give a hardness of approximation result which
  matches up to constant factors.  We also discuss negative results
  for simpler approaches and present encouraging early experimental
  results.
\end{abstract}

\section{Introduction}
As a motivating example, we consider viral marketing in a social
network.  In the standard version of the problem, the goal is to send
advertisements to influential members of a social network such that by
sending advertisements to only a few people our message spreads to a
large portion of the network.  Previous work \cite{influentialnodes,
  maximizingspread} has shown that, for many models of influence, the
influence of a set of nodes can be modelled as a submodular set
function.  Therefore, selecting a small set of nodes with maximal
influence can be posed as a \emph{submodular function maximization}
problem.  The related problem of selecting a minimal set of nodes to
achieve a desired influence is a \emph{submodular set cover} problem.
Both of these problems can be approximately solved via a simple greedy
approximation algorithm.

Consider a variation of this problem in which the goal is not to send
advertisements to people that are influential in the entire social
network but rather to people that are influential in a specific target
group.  For example, our target group could be people that like
snowboarding or people that listen to jazz music.  If the members of
the target group are unknown and we have no way of learning the
members of the target group, there is little we can do except assume
every member of the social network is a member of the target group.
However, if we assume the group has some known structure and that we
receive feedback from sending advertisements (e.g.\ in the form of ad
clicks or survey responses), it may be possible to simultaneously
discover the members of the group and find people that are influential
in the group.

We call problems like this \emph{learning and covering problems}.  In
our example, the \emph{learning} aspect of the problem is discovering
the members of the target group (the people that like snowboarding),
and the \emph{covering} aspect of the problem is to select a small set
of people that achieve a desired level of influence in the target
group (the people to target with advertisements).  Other applications
have similar structure.  For example, we may want to select a small
set of representative documents about a topic of interest (e.g.\ about
linear algebra).  If we do not initially know the topic labels for
documents, this is also a learning and covering problem.

We propose a new problem called \emph{interactive submodular set
  cover} that can be used to model many learning and covering
problems.  Besides addressing interesting new applications,
interactive submodular set cover directly generalizes submodular set
cover and exact active learning with a finite hypothesis class (query
learning) giving new insight into many previous theoretical results.
We derive and analyze a new algorithm that is guaranteed to perform
approximately as well as any other algorithm and in fact has the best
possible approximation ratio.  Our algorithm considers simultaneously 
the learning and covering parts of the problem.  It is tempting to try
to treat these two parts of the problem separately for example by
first solving the learning problem and then solving the covering
problem.  We prove this approach and other simple approaches may
perform much worse than the optimal algorithm.

\section{Background}

\subsection{Submodular Set Cover}

A submodular function is a set function satisfying a natural
diminishing returns property.  We call a set function $F$ defined
over a ground set $V$ submodular iff for all $A \subseteq B \subseteq V$
and $v \in V \setminus B$
\begin{equation}
F(A + v) - F(A) \geq F(B + v) - F(B) 
\label{submodeq}
\end{equation}
In other words, adding an element to $A$, a subset of $B$, results in
a larger gain than adding the same element to $B$.  $F$ is called
modular if Equation \ref{submodeq} holds with equality.  $F$ is
monotone non-decreasing if for all $A \subseteq B \subseteq V$,
$F(A) \leq F(B)$.  Note that if $F$ is monotone non-decreasing
and submodular iff Equation \ref{submodeq} holds for all $v \in V$ 
(including $v \in B$).

\begin{prp} If $F_1(S), F_2(S), ... F_n(S)$ are all submodular, monotone
  non-decreasing functions then $F_1(S) + F_2(S) + ... + F_n(S)$ is
submodular, monotone non-decreasing. \label{sumprp} \end{prp}

\begin{prp} For any function $f$ mapping set elements to real numbers
the function $F(S) \triangleq \max_{s \in S} f(s)$ is a submodular, monotone
non-deceasing function. \label{maxprp} \end{prp}

In the submodular set cover problem the goal is to find a set $S
\subseteq V$ minimizing a modular cost function $c(S) = \sum_{s \in S}
c(s)$ subject to the constraint $F(S)=F(V)$ for a monotone
non-decreasing submodular $F$.  

\begin{leftbar}
\noindent \textbf{\large{Submodular Set Cover}} \\
\textbf{Given:} \ 
\begin{compactitem} 
\item Ground set $V$ 
\item Modular cost function $c$ defined over $V$
\item Submodular monotone non-decreasing 
objective function $F$ defined over $V$ 
\end{compactitem}
\textbf{Objective:} Minimize $c(S)$ such that $F(S) = F(V)$ 
\end{leftbar}

This problem is closely related to the problem of submodular function
maximization under a modular cost constraint $c(S) < k$ for a constant
$k$.  A number of interesting real world applications can be posed as
submodular set cover or submodular function maximization problems
including influence maximization in social networks
\cite{maximizingspread}, sensor placement and experiment design
\cite{robust}, and document summarization \cite{documentsum}.  In the
sensor placement problem, for example, the ground set $V$ corresponds
to a set of possible locations.  An objective function $F(S)$ measures
the coverage achieved by deploying sensors to the locations
corresponding to $S \subseteq V$.  For many reasonable definitions of
coverage, $F(S)$ turns out to be submodular.

Submodular set cover is a generalization of the set cover problem.  In
particular, set cover corresponds to the case where each $v \in V$ is
a set of items taken from a set $\bigcup_{v \in V} v$.  The goal is to
find a small set of sets $S \subseteq V$ such that $|\bigcup_{s \in S}
s| = |\bigcup_{v \in V} v|$.  The function $F(S) = |\bigcup_{s \in S}
s|$ is monotone non-decreasing and submodular, so this is a submodular
set cover problem.  As is the case for set cover, a greedy algorithm
has approximation guarantees for submodular set cover
\cite{analysisofgreedy}.  In particular, if $F$ is integer valued,
then the greedy solution is within $H(\max_{v \in V} F(\lbrace v
\rbrace))$ of the optimal solution where $H(k)$ is the $k$th harmonic
number.  Up to lower order terms, this matches the hardness of
approximation lower bound $(1 - o(1))\ln n$ \cite{thresholdsetcover}
where $n = |\bigcup_{v \in V} v| = F(V)$.

We note a variation of submodular set cover uses a constraint $F(S)
\geq \alpha$ for a fixed threshold $\alpha$.  This variation does not
add any difficulty to the problem because we can always define a new
monotone non-decreasing submodular function $\hat{F}(S) = \min(F(S),
\alpha)$ \cite{robust, electrical} to convert the constraint $F(S) \geq
\alpha$ into a new constraint $\hat{F}(S) = \hat{F}(V)$.  We can also
convert in the other direction from a constraint $F(S) = F(V)$ to
$F(S) \geq \alpha$ by setting $\alpha = F(V)$.  Without loss of
generality or specificity, we use the variation of the problem with an
explicit threshold $F(S) \geq \alpha$.

\subsection{Exact Active Learning}

In the exact active learning problem we have a known finite hypothesis
class given by a set of objects $H$, and we want to identify an
initially unknown target hypothesis $h^* \in H$.  We identify $h^*$ by
asking questions.  Define $Q$ to be the known set of all possible
questions.  A question $q$ maps an object $h$ to a set of valid
responses $q(h) \subseteq R$ with $q(h) \neq \emptyset$ where $R
\triangleq \bigcup_{q \in Q, h \in H} q(h)$ is the set of all possible
responses.  We know the mapping for each $q$ (i.e.\ we know $q(h)$ for
every $q$ and $h$).  Asking $q$ reveals some element $r \in q(h^*)$
which may be chosen adversarially (chosen to impede the learning
algorithm).  Each question $q \in Q$ has a positive cost $c(q)$ defined
by the modular cost function $c$.

The goal of active learning is to ask a sequence of questions with
small total cost that identifies $h^*$.  By identifying $h^*$, we mean
that for every $h \neq h^*$ we have received some response $r$ to a
question $q$ such that $r \notin q(h)$.  Questions are chosen
sequentially so that the response from a previous question can be used
to decide which question to ask next.  The problem is stated below.

\begin{leftbar}
\noindent \textbf{\large{Exact Active Learning}} \\
\textbf{Given:} \ 
\begin{compactitem}
\item Hypothesis class $H$ containing an unknown target $h^*$
\item Query set $Q$ and response set $R$ 
with $q(h) \subseteq R$ for $q \in Q$, $h \in H$ 
\item Modular query cost function $c$ defined over $Q$
\end{compactitem}
\textbf{Repeat:} Ask a question $\hat{q}_i \in Q$ and receive
a response $\hat{r}_i \in \hat{q}_i(h^*)$ \\
\textbf{Until:} $h^*$ is identified (for every $h \in H$ with $h \neq
h^*$ there is a $(\hat{q}_i, \hat{r}_i)$ with $\hat{r}_i \notin \hat{q}_i(h)$) \\
\textbf{Objective:} Minimize $\sum_i c(\hat{q}_i)$
\end{leftbar}

In a typical exact learning problem, $H$ is a set of different classifiers
and $h^*$ is a unique zero-error classifier.  Questions in $Q$ can, for
example, correspond to label (membership) queries for data points.  If
we have a fixed data set consisting of data points $x_i$, we can create
a question $q_i$ corresponding to each $x_i$ and set $q_i(h) = \lbrace
h(x_i) \rbrace$.  Questions can also correspond to more complicated
queries.  For example, a question can ask if any points in a set are
positively labelled.  The setting we have described allows for
mixing arbitrary types of queries with different costs.  

For a set of question-response pairs $\hat{S}$, define
the version space $V(\hat{S})$ to be the subset of $H$ consistent
with $\hat{S}$
\[ V(\hat{S}) \triangleq 
\lbrace h \in H : \forall (q, r) \in \hat{S}, \ r \in q(h) \rbrace \]
In terms of the version space, the goal of exact active learning
is to ask a sequence of questions such that $|V(\hat{S})| = 1$.

We note that the assumption that $H$ and $Q$ are finite is not a
problem for many applications involving finite data sets.  In
particular, if we have an infinite a hypothesis class (e.g.\ linear
classifiers with dimension $d$) and a finite data set, we can simply
use the effective hypothesis class induced by the data set
\cite{analysisofgreedyactive}.  On the other hand, the assumption that
we have direct access to the target hypothesis (every $\hat{r}_i$ is in
$\hat{q}_i(h^*)$) and that the target hypothesis is in our hypothesis class
($h^* \in H$) is a limiting assumption.  Stated differently, we assume
that there is no noise and that the hypothesis class is correct.

Building on previous work \cite{generaldim}, \citet{costcomp} showed
that a simple greedy active learning strategy is approximately optimal
in the setting we have described.  The greedy strategy selects the
question which relative to cost distinguishes the greatest number of
hypotheses from $h^*$. \citet{costcomp} shows this strategy incurs no
more than $\ln |H|$ times the cost of any other question asking
strategy.

The algorithms and approximation factors for submodular set cover and
exact active learning are quite similar.  Both are simple greedy
algorithms and the $\ln F(V)$ approximation for submodular set cover
is similar to the $\ln |H|$ approximation for active learning.  These
similarities suggest these problems may be special cases of some other
more general problem.  We show that in fact they are special cases of
a problem which we call interactive submodular set cover.

\section{Problem Statement}

We use notation similar to the exact active learning problem we
described in the previous section.  Assume we have a finite hypothesis
class $H$ containing an unknown target hypothesis $h^* \in H$.  We
again assume there is a finite set of questions $Q$, a question $q$
maps each object $h$ to a set of valid responses $q(h) \subseteq R$
with $q(h) \neq \emptyset$, and each question $q \in Q$ has a positive
cost $c(q)$ defined by the modular cost function $c$.  We also again assume
that we know the mapping for each $q$ (i.e.\ we know $q(h)$ for every
$q$ and $h$).  Asking $q$ reveals some adversarially chosen element $r
\in q(h^*)$.  In the exact active learning problem the goal is to
identify $h^*$ through questions.  In this work we consider a
generalization of this problem in which the goal is instead to satisfy
a submodular constraint that depends on $h^*$.  

We assume that for each object $h$ there is a corresponding monotone
non-decreasing submodular function $F_h$ defined over subsets of $Q
\times R$ (sets of question-response pairs).  We repeatedly ask a
question $\hat q_i$ and receive a response $\hat r_i$. Let the
sequence of questions be $\hat Q = (\hat q_1, \hat q_2, \dots)$
and sequence of responses be $\hat R = (\hat r_1, \hat r_2, \dots)$.
Define $\hat{S} = \bigcup_{\hat{q}_i \in \hat{Q}} \lbrace (\hat{q}_i,
\hat{r}_i) \rbrace$ to be the final set of question-response pairs
corresponding to these sequences.  Our goal is to ask a sequence of
questions with minimal total cost $c(\hat{Q})$ which ensures
$F_{h^*}(\hat{S}) \geq \alpha$ for some threshold $\alpha$ without
knowing $h^*$ beforehand.  We call this problem interactive submodular
set cover.

\begin{leftbar}
\noindent \textbf{\large{Interactive Submodular Set Cover}} \\
\textbf{Given:} \ 
\begin{compactitem}
\item Hypothesis class $H$ containing an unknown target $h^*$
\item Query set $Q$ and response set $R$ 
with known $q(h) \subseteq R$ for every $q \in Q$, $h \in H$ 
\item Modular query cost function $c$ defined over $Q$
\item Submodular monotone non-decreasing 
objective functions $F_h$ for $h \in H$ defined over $Q \times R$
\item Objective threshold $\alpha$
\end{compactitem}
\textbf{Repeat:} Ask a question $\hat{q}_i \in Q$ and receive
a response $\hat{r}_i \in \hat{q}_i(h^*)$ \\
\textbf{Until:} $F_{h^*}(\hat{S}) \geq \alpha$ 
where $\hat{S} = \bigcup_i \lbrace (\hat{q}_i, \hat{r}_i) \rbrace$  \\
\textbf{Objective:} Minimize $\sum_i c(\hat{q}_i)$ 
\end{leftbar}

Note  that  although  we  know   the  hypothesis  class  $H$ and the
corresponding  objective functions  $F_h$,  we do  not initially  know
$h^*$.  Information about  $h^*$ is only revealed as  we ask questions
and receive  responses to questions.  Responses  to previous questions
can be  used to  decide which question to ask  next, so in this way the
problem is  ``interactive.''  Furthermore, the  objective function for
each hypothesis $F_h$ is  defined over sets of question-response pairs
(as opposed to, say, sets of questions), so when asking a new question
we cannot  predict how the value  of $F_h$ will change  until after we
receive a response.   The only restriction on the response we receive
is that it must be consistent with the initially unknown target $h^*$.
It is  this uncertainty about $h^*$  and the feedback  we receive from
questions that distinguishes the problem from submodular set cover and
allows us to model learning and covering problems.

\subsection{Connection to Submodular Set Cover}

If we know $h^*$ (e.g.\ if $|H|=1$) and we assume $|q(h)|=1 $ $\forall
q \in Q, h \in H$ (i.e.\ that there is only one valid response to
every question), our problem reduces exactly to the standard
submodular set cover problem.  Under these assumptions, we can compute
$F_{h^*}(\hat{S})$ for any set of questions without actually asking
these questions.  \citet{robust} study a non-interactive version of
interactive submodular set cover in which $|q(h)|=1$ $\forall q \in Q,
h \in H$ and the entire sequence of questions must be chosen before
receiving any responses.  This restricted version of the problem can
also be reduced to standard submodular set cover \citet{robust}.

\subsection{Connection to Active Learning}

\label{connectiontoactivesec}

Define
\begin{equation*}
 F_{h}(\hat{S}) \triangleq F(\hat{S}) = |H \setminus V(\hat{S})| 
\end{equation*}
where $V(\hat{S})$ is again the version space (the set of hypotheses
consistent with $\hat{S}$).  This objective is the number of hypotheses
eliminated from the version space by $\hat{S}$.

\begin{lma} $F_{h}(\hat{S}) \triangleq |H \setminus V(\hat{S})|$
is submodular and monotone non-decreasing \label{vslma} \end{lma}

\begin{proof} To see this note that we can write $F_h$ as
$F_h(\hat{S}) = \sum_{h' \in H} \max_{(q, r) \in \hat{S}} f_{h'}((q, r))$
where $f_{h'}((q, r)) = 1$ if $r \notin q(h')$ and else
$f_{h'}((q, r)) = 0$.  The result then follows
from Proposition \ref{sumprp} and Proposition \ref{maxprp}. \end{proof}

For this objective, if we set $\alpha = |H| - 1$ we get the standard
exact active learning problem: our goal is to identify $h^*$ using a
set of questions with small total cost.  Note that in this case the
objective $F_h$ does not actually depend on $h$ (i.e.  $F_h = F_{h'}$
for all $h, h' \in H$) but the problem still differs from standard
submodular set cover because $F_h(\hat{S})$ is defined over
question-response pairs.

Interactive submodular set cover can also model an approximate
variation of active learning with a finite hypothesis class and finite
data set.  Define 
\[ F_{h}(\hat{S}) \triangleq |H \setminus V(\hat{S})| (|X| - \kappa)
+ \sum_{h' \in V(\hat{S})} \min(|X| - \kappa, \sum_{x \in X} I(h'(x) = h(x))) \] 
where $I$ is the indicator function, $X$ is a finite data set, and
$\kappa$ is an integer.  

\begin{prp}$F_{h^*}(\hat{S}) = |H| (|X| - \kappa)$ iff all
hypotheses in the version space make at most $\kappa$ mistakes.\end{prp}  

\begin{lma} 
$F_{h}(\hat{S}) \triangleq |H \setminus V(\hat{S})| (|X| - \kappa)
+ \sum_{h' \in V(\hat{S})} \min(|X| - \kappa, \sum_{x \in X} I(h'(x) = h(x)))$
is submodular and monotone non-decreasing \end{lma}

\begin{proof} We can write $F_h$ as
$F_h(\hat{S}) = \sum_{h' \in H} \max_{(q, r) \in \hat{S}} f_h((q, r))$
where $f_{h'}((q, r)) = |X| - \kappa$ if $r \notin q(h')$
and else $f_{h'}((q, r))=  \min(|X| - \kappa, \sum_{x \in X} I(h'(x) = h(x)))$.  
The result then follows from Proposition \ref{sumprp} and 
Proposition \ref{maxprp}. \end{proof}

For this objective, if we set $\alpha = |H|(|X| - \kappa)$ then our
goal is to ask a sequence of questions such that all hypotheses in the
version space make at most $\kappa$ mistakes.  \citet{generaldim}
study a similar approximate query learning setting, and
\citet{querycollab} consider a slightly different setting where the
target hypothesis may not be in $H$.

\subsection{Connection to Adaptive Submodularity}

In concurrent work, \citet{adaptsubmod} show results similar to ours
for a different but related class of problems which also involve
interactive (i.e.\ sequential, adaptive) optimization of submodular
functions.  What \citeauthor{adaptsubmod} call realizations correspond
to hypotheses in our work while items and states correspond to queries
and responses respectively.  \citeauthor{adaptsubmod} consider both
average-case and worst-case settings and both maximization and
min-cost coverage problems.  In contrast, we only consider worst-case,
min-cost coverage problems.  In this sense our results are less
general.

However, in other ways our results are more general.  The main greedy
approximation guarantees shown by \citeauthor{adaptsubmod} require
that the problem is \emph{adaptive submodular}; adaptive submodularity
depends not only on the objective but also on the set of possible
realizations and the probability distribution over these realizations.
In contrast we only require that for a fixed hypothesis the objective
is submodular.  \citeauthor{adaptsubmod} call this pointwise
submodularity.  Pointwise submodularity does not in general imply
adaptive submodularity (see the clustered failure model discussed by
\citeauthor{adaptsubmod}).  

In fact, for problems that are pointwise modular but not adaptive
submodular, \citeauthor{adaptsubmod} show a hardness of approximation
lower bound of $\mathcal{O}(|Q|^{1-\epsilon})$; we note this does not
contradict our results as their proof is for average-case cost and uses
a hypothesis class with $|H|=2^{|Q|}$.  \citeauthor{adaptsubmod} also
propose a simple non greedy approach with explicit explore and exploit
stages; this approach requires only a weaker assumption that the value
of the exploitation stage is adaptive submodular with respect to
exploration.  However, it is not immediately obvious when this
condition holds, and it is also not clear how to apply this approach
to worst-case or min-cost coverage problems.

There are other smaller differences between our problem settings: we
let queries map hypotheses to \emph{sets} of valid responses (in
general $|q(h)|>1$) while \citeauthor{adaptsubmod} define realizations
as maps from items to \emph{single} states.  Also, in our work we
allow for non uniform query costs (in general $c(q_i) \neq c(q_j)$)
while \citeauthor{adaptsubmod} require that every item has the same
cost (\citeauthor{adaptsubmod} do however mention that the extension
to non uniform costs is straightforward).  We finally note that the
proof techniques we use are quite different.

Some other previous work has also considered interactive versions of
covering problems in an average-case model \cite{maximizingstochastic,
  stochasticcovering}.  The work of \citet{maximizingstochastic} is
perhaps most similar and considers a submodular function maximization
problem over independent random variables which are sequentially
queried.  The setting considered by \citet{adaptsubmod} strictly
generalizes this setting.  \citet{onlinemaximizing} study an online
version of submodular function maximization where a sequence of
submodular function maximization problems is solved.  This problem is
related in that it also involves learning and submodular functions,
but the setting is very different than the one studied here where we
solve a single interactive problem as opposed to a series of
non-interactive problems.

\section{Example}

\label{examplesec}

\begin{figure}[t]
\begin{center}
\includegraphics[width=.2\columnwidth]{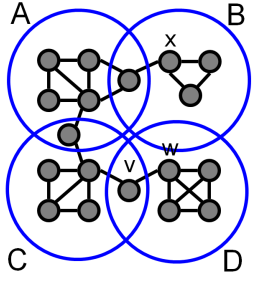}
\end{center}
\caption{A cartoon example social network.}
\label{cartoon}
\end{figure}

In the advertising application we described in the introduction, the
target hypothesis $h^*$ corresponds to the group of people we want to
target with advertisements (e.g.\ the people that like snowboarding),
and the hypothesis class $H$ encodes our prior knowledge about $h^*$.
For example, if we know the target group forms a small dense subgraph
in the social network, then the hypothesis class $H$ would be the set
of \emph{all} small dense subgraphs in the social network.  The query
set $Q$ and response set $R$ correspond to advertising actions and
feedback respectively, and finally the objective function $F_h$
measures advertising coverage within the group corresponding to $h$.

To make the discussion concrete, assume the advertiser sends a single
ad at a time and that after a person is sent an ad the advertiser
receives a binary response indicating if that person is in the target
group (i.e.\ likes snowboarding).  Let $q_i$ correspond to sending an
ad to user $i$ (i.e.\ node $i$), and $q_i(h) = \lbrace 1 \rbrace$ if
user $i$ is in group $h$ and $q_i(h) = \lbrace 0 \rbrace$ otherwise.
For our coverage goal, assume the advertiser wants to ensure that
every person in the target group either receives an ad or has a friend
that receives an ad.  We say a node is ``covered'' if it has received
an ad or has a neighbor that has received an ad.  This is a variation of the
minimum dominating set problem, and we use the following objective
\[F_h(\hat{S}) \triangleq \sum_{v \in V_h} I \Big( v \in V_{\hat{S}} \mbox{ or }
\exists s \in V_{\hat{S}} : (v, s) \in E \Big) + |V \setminus V_h| \] where
$V$ and $E$ are the nodes and edges in the social network, $V_h$ is
the set of nodes in group $h$, and $V_{\hat{S}}$ is the set of nodes
corresponding to ads we have sent.  With this objective
$F_{h^*}(\hat{S}) = |V|$ iff we have achieved the stated coverage goal.

\begin{lma} 
$F_h(\hat{S}) = \sum_{v \in V_h} I ( v \in V_{\hat{S}} \mbox{ or }
\exists s \in V_{\hat{S}} : (v, s) \in E ) + |V \setminus V_h| $
is submodular and monotone non-decreasing. 
\end{lma}

\begin{proof} We can write $F_h(\hat{S})$ as 
$F_h(\hat{S}) = \sum_{v \in V} \max_{(\hat{q}, \hat{r}) \in \hat{S}} 
f_v((\hat{q}, \hat{r}))$ where 
$f_v((\hat{q}, \hat{r})) = 1$ if the action $\hat{q}$ 
covers $v$ or $v \notin V_h$ and
$f_v((\hat{q}, \hat{r})) = 0$ otherwise.  The result then follows
from Proposition \ref{sumprp} and Proposition \ref{maxprp}.  \end{proof}

Figure \ref{cartoon} shows a cartoon social network.  For this
example, assume the advertiser knows the target group is one of the
four clusters shown (marked A, B, C, and D) but does not know which.
This is our hypothesis class $H$.  The node marked $v$ is initially
very useful for learning the members of the target group: if we send
an ad to this node, no matter what response we receive we are
guaranteed to eliminate two of the four clusters (either $A$ and $B$
or $C$ and $D$).  However, this node has only a degree of 2 and
therefore sending an ad to this node does not cover very many nodes.
On the other hand, the nodes marked $x$ and $w$ are connected to every
node in clusters $B$ and $D$ respectively.  $x$ (resp.\ $w$) is
therefore very useful for achieving the coverage objective if the
target group is $B$ (resp.\ $D$).  An algorithm for learning and
covering must choose between actions more beneficial for learning vs.\
actions more beneficial for covering (although sometimes an action can
be beneficial for both to a certain degree). The interplay between
learning and covering is similar to the exploration-exploitation
trade-off in reinforcement learning.  In this example an optimal
strategy is to first send an ad to $v$ and then cover the remaining
two clusters using two additional ads for a worst case cost of 3.

A simple approach to learning and covering is to simply ignore
feedback and solve the covering problem for all possible target
groups.  In our example application the resulting covering problem is
a simple dominating set problem for which we can use standard
submodular set cover methods.  We call this the Cover All strategy.
This approach is suboptimal because in many cases feedback can make
the problem significantly easier.  In our synthetic example, any
strategy not using feedback must use worst case cost of 4: four
ads are required to cover all of the nodes in the four clusters.  Theorem
\ref{coverallgap} in 
Section \ref{negresultsec} proves that in fact there are cases where
the best strategy not using feedback incurs exponentially greater cost
than the best strategy using feedback.

Another simple approach is to solve the learning problem first
(identify $h^*$) and then solve the covering problem (satisfy
$F_{h^*}(\hat{S})$).  We can use, for example, query learning to solve
the learning problem and then use standard submodular set cover to
solve the covering problem.  We call this the Learn then Cover
strategy.  This approach turns out to match the optimal strategy in
the example given by Figure \ref{cartoon}.  In this example the target
group can be identified using 2 queries by querying $v$ then $w$ if
the response is $1$ and $x$ if the response is $0$.  After identifying
the target group, the target group can be covered in at most one
more query.  However, this approach is not optimal for other instances
of this problem.  For example, if we were to add an additional node
which is connected to every other node then the covering problem would
have a solution of cost 1 while the learning problem would still
require cost of 2.  Theorem \ref{learnthencovergap} in Section
\ref{negresultsec} shows that in fact there are examples where solving
a learning problem is much harder than solving the corresponding
learning and covering problem.  We therefore must consider other
methods for balancing learning and covering.

We note that this problem setup can be modified to allow queries to
have sometimes uninformative responses; this can be modeled by adding
an additional response to $R$ which corresponds to a ``no-feedback''
response and including this response in the set of allowable responses
($q(h)$) for certain query-hypothesis pairs .  However, care must be
taken to ensure that the resulting problem is still interesting for
worst-case choice of responses; if we allow ``no-feedback'' responses
for every question-hypothesis pair, then the in the worst-case we will
never receive any feedback, so a worst case optimal strategy could
ignore all responses.

\section{Greedy Approximation Guarantee}

We are interested in approximately optimal polynomial time algorithms
for the interactive submodular set cover problem.  We call a question
asking strategy correct if it always asks a sequence of questions such
that $F_{h^*}(\hat{S}) \geq \alpha$ where $\hat{S}$ is again the final
set of question-response pairs.  A necessary and sufficient condition
to ensure $F_{h^*}(\hat{S}) \geq \alpha$ for worst case choice of
$h^*$ is to ensure $\min_{h \in V(\hat{S})} F_{h}(\hat{S}) \geq
\alpha$ where $V(\hat{S})$ is the version space.  Then a simple
stopping condition which ensures a question asking strategy is correct
is to continue asking questions until $\min_{h \in V(\hat{S})}
F_{h}(\hat{S}) \geq \alpha$.  We call a question asking strategy
approximately optimal if it is correct and the worst case cost
incurred by the strategy is not much worse than the worst case cost of
any other strategy.

As discussed informally in the previous section, it is important for a
question asking strategy to balance between learning (identifying
$h^*$) and covering (increasing $F_{h^*}$).  Ignoring either aspect of
the problem is in general suboptimal (we show this formally in Section
\ref{negresultsec}).  We propose a reduction which converts the
problem over many objective functions $F_h$ into a problem over a
single objective function $\bar{F}_{\alpha}$ that encodes the trade-off
between learning and covering.  We can then use a greedy algorithm to
maximize this single objective, and this turns out to overcome the
shortcomings of simpler approaches.  This reduction is inspired by the
reduction used by \citet{robust} in the non-interactive setting to
convert multiple covering constraints into a single covering
constraint.

Define
\[ \bar{F}_{\alpha}(\hat{S}) \triangleq (1/|H|) (\sum_{h \in
  V(\hat{S})} \min(\alpha, F_h(\hat{S})) + \alpha |H \setminus
V(\hat{S})|) \] $\bar{F}_{\alpha} (\hat{S}) \geq \alpha$ iff
$F_h(\hat{S}) \geq \alpha$ for all $h \in V(\hat{S})$ so a question
asking strategy is correct iff it satisfies $\bar{F}_{\alpha}
(\hat{S}) \geq \alpha$.  This objective balances the value of learning
and covering.  The sum over $h \in V(\hat{S})$ measures progress
towards satisfying the covering constraint for hypotheses $h$ in the current
version space (covering).  The second term $\alpha |H \setminus
V(\hat{S})|$ measures progress towards identifying $h^*$ through reduction in
version space size (learning).
Note that the objective does not make a hard distinction between
learning actions and covering actions.  In fact, the objective will
prefer actions that both increase $F_h(\hat{S})$ for $h \in V(\hat{S})$
and decrease the size of $V(\hat{S})$.  Crucially, $\bar{F}_{\alpha}$
retains submodularity.

\begin{lma}
$\bar{F}_{\alpha}$ is submodular and monotone non-decreasing
when every $F_h$ is submodular and monotone non-decreasing.
\end{lma}

\begin{proof} Note that the proof would be trivial if the sum
were over all $h \in H$.  However, since the sum is over a subset of
$H$ which depends on $\hat{S}$, the result is not obvious.  
We can write $\bar{F}_{\alpha}$ as 
$\bar{F}_{\alpha}(\hat{S}) = (1/|H|) \sum_{h \in H} \hat{F}_{\alpha,  h}(\hat{S})$ 
where we define $\hat{F}_{\alpha,  h}(\hat{S}) \triangleq 
I(h \in V(\hat{S})) \min(\alpha, F_h(\hat{S})) 
+ I(h \notin V(\hat{S})) \alpha$. It is not hard to 
see $\hat{F}_{\alpha,  h}$ is monotone non-decreasing.  
We show $\hat{F}_{\alpha,  h}$ is also submodular and 
the result then follows from Proposition \ref{sumprp}.  Consider any
$(q, r) \notin B$ and $A \subseteq B \subseteq (Q \times R)$.  We show
Equation \ref{submodeq} holds in three cases.  Here we use
as short hand $\Gain(F, S, s) \triangleq F(S + s) - F(S)$. 
\begin{compactitem}
\item If \textbf{$h \notin V(B)$} then
\[ \Gain(\hat{F}_{\alpha,  h},A,(q, r)) 
\geq 0 = \Gain(\hat{F}_{\alpha,  h},B,(q, r)) \]
\item If \textbf{$r \notin q(h)$} then
\[ \Gain(\hat{F}_{\alpha,  h},A,(q, r))
 = \alpha - \hat{F}_{\alpha,  h}(A) 
 \geq \alpha -  \hat{F}_{\alpha,  h}(B) 
 = \Gain(\hat{F}_{\alpha,  h},B,(q, r))  \]
\item If \textbf{$r \in q(h)$ and $h \in V(B)$} then
\begin{align*}
\Gain(\hat{F}_{\alpha,  h},A,(q, r)) 
& = \min(F_h(A + (q, r)), \alpha) - \min(F_h(A), \alpha) \\
&  \geq \min(F_h(B + (q, r)), \alpha) - \min(F_h(B), \alpha)
 = \Gain(\hat{F}_{\alpha,  h},B,(q, r)) 
\end{align*}
Here we used the submodularity of $\min(F_h(S), \alpha)$ \cite{electrical}.
\end{compactitem}
\end{proof}

\begin{algorithm}[t]
\caption{Worst Case Greedy}
\begin{algorithmic}[1]
\STATE $\hat{H} \Leftarrow H$
\STATE $\hat{S} \Leftarrow \emptyset$
\WHILE{$\bar{F}_{\alpha} (\hat{S}) < \alpha$}
\STATE $\hat{q} \Leftarrow \operatorname{argmax}_{q_i \in Q} 
\min_{h \in V(\hat{S})} \min_{r_i \in q_i(h)} 
(\bar{F}_{\alpha}(\hat{S} + (q_i, r_i)) - 
\bar{F}_{\alpha}(\hat{S})) / c(q_i) $
\STATE Ask $\hat{q}$ and receive response $\hat{r}$
\STATE $\hat{S} \Leftarrow \hat{S} + (\hat{q}, \hat{r})$
\ENDWHILE
\end{algorithmic}
\label{greedyb}
\end{algorithm}

Algorithm \ref{greedyb} shows the worst case greedy algorithm which
at each step picks the question $q_i$ that maximizes the worst
case gain of $\bar{F}_{\alpha}$ 
\[ \min_{h \in V(\hat{S})} \min_{r_i \in q_i(h)}
(\bar{F}_{\alpha}(\hat{S} + (q_i, r_i)) - \bar{F}_{\alpha}(\hat{S})) /
c(q_i) \] We now argue that Algorithm \ref{greedyb} is an
approximately optimal algorithm for interactive submodular set cover.
Note that although it is a simple greedy algorithm over a single
submodular objective, the standard submodular set cover analysis
doesn't apply: the objective function is defined over
question-response pairs, and the algorithm cannot predict the actual
objective function gain until after selecting and commiting to a
question and receiving a response.  We use an Extended Teaching
Dimension style analysis \cite{costcomp} inspired by previous work in
query learning.  We are the first to our knowledge to use this kind of
proof for a submodular optimization problem.

Define an oracle (teacher) $T \in R^Q$ to be a function mapping 
questions to responses.  As a short hand, for a sequence of 
questions $\hat{Q}$ define 
\[ 
T(\hat{Q}) \triangleq \bigcup_{\hat{q}_i \in \hat{Q}} 
\lbrace (\hat{q}_i, T(\hat{q}_i)) \rbrace 
\]
$T(\hat{Q})$ is the set of question-response pairs received when $T$
is used to answer the questions in $\hat{Q}$.  We now define a
quantity analogous to the General Identification Cost for exact active
learning \cite{costcomp}.  Define the General Cover Cost, $GCC$
\[ GCC \triangleq \max_{T \in R^Q} (\min_{\hat{Q} :
  \bar{F}_{\alpha}(T(\hat{Q})) \geq \alpha} c(\hat{Q})) \] $GCC$
depends on $H$, $Q$, $\alpha$, $c$, and the objective functions $F_h$,
but for simplicity of notation this dependence is suppressed.  $GCC$
can be viewed as the cost of satisfying $\bar{F}_{\alpha}(T(\hat{Q}))
\geq \alpha$ for worst case choice of $T$ where the choice of $T$ is
\emph{known to the algorithm selecting $\hat{Q}$}.  Here the worst
case choice of $T$ is over all mappings between $Q$ and $R$.  There is
no restriction that $T$ answer questions in a manner consistent with
any hypothesis $h \in H$.

We first show that $GCC$ is a lower bound on the optimal worst case cost
of satisfying $F_{h^*}(\hat{S}) \geq \alpha$.

\begin{lma}If there is a correct question asking strategy for
satisfying $F_{h^*}(\hat{S}) \geq \alpha$ with worst
case cost $C^*$ then $GCC \leq C^*$. \label{lowerlma} \end{lma}

\begin{proof}Assume the lemma is false and there is a correct question
asking strategy with worst case cost $C^*$ and $GCC > C^*$.  Using
this assumption and the definition of $GCC$, there is some oracle
$T^*$ such that
\[ \min_{\hat{Q} : \bar{F}_{\alpha}(T^*(\hat{Q})) \geq \alpha}
c(\hat{Q}) = GCC > C^* \]
When we use $T^*$ to answer questions, any sequence of
questions $\hat{Q}$ with total cost less than or equal to $C^*$ 
must have $\bar{F}_{\alpha}(\hat{S}) < \alpha$.  
$\bar{F}_{\alpha}(\hat{S}) < \alpha$ in turn 
implies $F_{h^*}(\hat{S}) < \alpha$ for some target hypothesis choice
$h^* \in V(\hat{S})$.  This contradicts
the assumption there is a correct strategy with worst
case cost $C^*$.\end{proof}

We now establish that when $GCC$ is small, there must be a question
which increases $\bar{F}_{\alpha}$.  
\begin{lma}For any initial set of questions-response pairs $\hat{S}$,
there must be a question $q \in Q$ such that
%\begin{multline*}
\[ \min_{h \in V(\hat{S})} \min_{r \in q(h)} 
\bar{F}_{\alpha}(\hat{S} + (q, r)) - 
\bar{F}_{\alpha}(\hat{S}) \geq 
 c(q) (\alpha - \bar{F}_{\alpha}(\hat{S})) / GCC \]
%\end{multline*}
\label{progresslma}
\end{lma}
\begin{proof}
Assume the lemma is false and for every question $q$ there is some
$h \in V(\hat{S})$ and $r \in q(h)$ such that
\[
\bar{F}_{\alpha}(\hat{S} + (q, r)) -
\bar{F}_{\alpha}(\hat{S}) < c(q) (\alpha - \bar{F}_{\alpha}(\hat{S})) / GCC
\]
Define an oracle $T'$ which answers every question with a response
satisfying this inequality.
For example, one such $T'$ is
\[ T'(q) \triangleq 
\operatorname{argmin}_{r} \bar{F}_{\alpha}(\hat{S} + (q, r)) 
-  \bar{F}_{\alpha}(\hat{S}) \]
By the definition of $GCC$
\[
\min_{\hat{Q} :
  \bar{F}_{\alpha}(T'(\hat{Q})) \geq \alpha} c(\hat{Q}))
\leq 
\max_{T \in R^Q} (\min_{\hat{Q} :
  \bar{F}_{\alpha}(T(\hat{Q})) \geq \alpha} c(\hat{Q})) = GCC 
\]
so there must be a sequence of questions 
$\hat{Q}$ with $c(\hat{Q}) \leq GCC$ such that
$\bar{F}_{\alpha}(T'(\hat{Q})) \geq \alpha$.
Because $\bar{F}_{\alpha}$ is monotone non-decreasing, we also know
$\bar{F}_{\alpha}(T'(\hat{Q}) \cup \hat{S}) \geq \alpha$.  Using the
submodularity of $\bar{F}_{\alpha}$,
\begin{eqnarray*}
%\lefteqn{\bar{F}_{\alpha}(T'(\hat{Q}) \cup \hat{S})} \\
\bar{F}_{\alpha}(T'(\hat{Q}) \cup \hat{S})
& \leq &  \bar{F}_{\alpha}(\hat{S}) + \sum_{q \in \hat{Q}} 
(\bar{F}_{\alpha}(\hat{S} \cup \lbrace (q, T(q)) \rbrace) - 
\bar{F}_{\alpha}(\hat{S})) \\
& < & \bar{F}_{\alpha}(\hat{S}) + \sum_{q \in \hat{Q}} 
c(q) (\alpha - \bar{F}_{\alpha}(\hat{S}))   / GCC \leq \alpha
\end{eqnarray*}
which is a contradiction.
\end{proof}

We can now show approximate optimality.
\begin{thm}
Assume that $\alpha$ is an integer and, for any $h \in H$, $F_h$ is an
integral monotone non-decreasing submodular function.  Algorithm
\ref{greedyb} incurs at most $GCC (1 + \ln (\alpha n))$ cost.
\label{mainthm}
\end{thm}

\begin{proof}
Let $\hat{q}_i$ be the question asked on the $i$th iteration, 
$\hat{S}_i$ be the set of question-response pairs after asking
$\hat{q}_i$ and $C_i$ be $\sum_{j \leq i} c(\hat{q}_j)$.  
By Lemma \ref{progresslma}
\[ \bar{F}_{\alpha}(\hat{S}_i) - \bar{F}_{\alpha}(\hat{S}_{i-1})
\geq c(\hat{q}_i) (\alpha - \bar{F}_{\alpha}(\hat{S}_{i-1})) / GCC \]
After some algebra we get
\[ \alpha - \bar{F}_{\alpha}(\hat{S}_i)
\leq  (\alpha - \bar{F}_{\alpha}(\hat{S}_{i-1})) (1 - c(\hat{q}_i) / GCC) \]
Now using $1-x < e^{-x}$
\[ \alpha - \bar{F}_{\alpha}(\hat{S}_i) \leq (\alpha -
\bar{F}_{\alpha}(\hat{S}_{i-1})) e^{-c(\hat{q}_i) / GCC} = \alpha
e^{-C_i / GCC} \] We have shown that the gap $\alpha -
\bar{F}_{\alpha}(\hat{S}_i)$ decreases exponentially fast with the
cost of the questions asked.  The remainder of the proof proceeds by
showing that (1) we can decrease the gap to $1/|H|$ using questions
with at most $GCC \ln (\alpha |H|)$ cost and (2) we can decrease the
gap from $1/|H|$ to $0$ with one question with cost at most $GCC$.

Let $j$ is the largest
integer such that
$\alpha - \bar{F}_{\alpha}(\hat{S}_j) \geq 1/|H|$ holds.  Then
\[
1/|H| \leq \alpha e^{-C_j / GCC}
\]
Solving for $C_j$ we get $C_j \leq GCC \ln (\alpha |H|)$.  This
completes (1).

By Lemma \ref{progresslma}, $\bar{F}_{\alpha}(\hat{S}_i) <
\bar{F}_{\alpha}(\hat{S}_{i+1})$ (we strictly increase the objective
on each iteration).  Because $\alpha$ is an integer and for every $h$
$F_h$ is an integral function, we can conclude
$\bar{F}_{\alpha}(\hat{S}_i) < \bar{F}_{\alpha}(\hat{S}_{i+1}) +
1/|H|$.  Then $q_{j+1}$ will be the final question asked.  By Lemma
\ref{progresslma}, $q_{j+1}$ can have cost no greater than $GCC$.
This completes (2).  We can finally conclude the cost incurred by the
greedy algorithm is at most $GCC (1 + \ln(\alpha |H|))$
\end{proof}

By combining Theorem \ref{mainthm} and Lemma \ref{lowerlma} we get
\begin{cor}
For integer $\alpha$ and integral monotone non-decreasing submodular $F_h$,
the worst case cost of Algorithm \ref{greedyb} is within
$1 + \ln (\alpha |H|)$ of that of any other correct question asking strategy
\end{cor}
We have shown a result for integer valued $\alpha$ and objective
functions.  We speculate that for more general non-integer objectives
it should be possible to give results similar to those for standard
submodular set cover \cite{analysisofgreedy}.  These approximation
bounds typically add an additional normalization term.

\section{Negative Results}

\label{negresultsec}

\subsection{\Naive Greedy}

The algorithm we propose is not the most obvious approach to the
problem.  A more direct extension of the standard submodular set cover
algorithm is to choose at each time step a question $q_i$ which has
not been asked before and that maximizes the worst case gain of
$F_{h^*}$.  In other words, chose the question $q_i$ that maximizes
\[ \min_{h \in V(\hat{S})} \min_{r_i \in q_i(h)} (F_h(\hat{S} + (q_i,
r_i)) - F_h(\hat{S})) / c(q_i) \] This is in contrast to the method we
propose that maximizes the worst-case gain of $\bar{F}_{\alpha}$
instead of $F_h$.  We call this strategy the \Naive Greedy Algorithm.
This algorithm in general performs much worse than the optimal
strategy.  The counter example is very similar to that given by
\citet{robust} for the equivalent approach in the non-interactive
setting.

\begin{thm}
Assume $F_h$ is integral for all $h \in H$ and $\alpha$ is integer. The 
\Naive Greedy Algorithm has approximation ratio at least
$\Omega(\alpha \max_i c(q_i) / \min_i c(q_i))$.  
\end{thm}

\begin{proof}
Consider the following example with $|H| = 2$, $|Q| = \alpha + 2$, $|R|=1$
and $\alpha>1$. When $|R|=1$ responses reveal no information about
$h^*$, so the interactive problem is equivalent to the non-interactive
problem, and the objective function only depends on the set of
questions asked.  Let $F_{h_1}$ and $F_{h_2}$ be modular 
functions defined by
\begin{align*}
F_{h_1}(q_1) \triangleq \alpha && 
F_{h_1}(q_2) \triangleq 0 \\
F_{h_2}(q_1) \triangleq 0 && 
F_{h_2}(q_2) \triangleq \alpha  
\end{align*}
and, for all $h$ and all $q_i$ with $i > 2$, $F_h(q_i) \triangleq 
1$.  The optimal strategy asks $q_1$ and $q_2$ (since $h^*$ is unknown
we must ask both).  However, the worst-case gain of asking $q_1$ or
$q_2$ is zero while the gain of asking $q_i$ for $i > 2$ is $1 /
c(q_i)$.  The \Naive Greedy Algorithm will then always ask every $q_i$
for $i > 2$ before asking $q_1$ and $q_2$ no matter how large $c(q_i)$
is compared to $c(q_1)$ and $c(q_2)$.  By making $c(q_i)$ for $i > 2$ 
large compared to $c(q_1)$ and $c(q_2)$ we get the claimed approximation
ratio.
\end{proof}

\subsection{Learn then Cover}

The method we propose for interactive submodular set cover
simultaneously solves the learning problem and covering problem in
parallel, only solving the learning problem to the extent that it
helps solve the covering problem.  A simpler strategy is to solve
these two problems in series (i.e.\  first identify $h^*$ using the
standard greedy query learning algorithm and second solve the
submodular set cover problem for $F_{h^*}$ using the standard greedy
set cover algorithm).  We call this the Learn then Cover approach.  We
show that this approach and in fact any approach that identifies $h^*$
exactly can perform very poorly.  Therefore it is important to
consider the learning problem and covering problem simultaneously.

\begin{thm}Assume $F_h$ is integer for all $h$ and 
that $\alpha$ is an integer.  Any algorithm that exactly identifies $h^*$
has approximation ratio at least $\Omega(|H| \max_i c(q_i) / \min_i c(q_i))$.
\label{learnthencovergap}
\end{thm}

\begin{proof}
We give a simple example for which the learning problem 
(identifying $h^*$) is hard
but the interactive submodular set cover problem (satisfying
$F_{h^*}(\hat{S}) \geq \alpha$) is easy.  
For $i \in {1 ... |H|}$ let
$q_i(h_j) = \lbrace 1 \rbrace$ if $i = j$ and 
$q_i(h_j) = \lbrace 0 \rbrace$ if $i \neq j$.  For $i = |H| + 1$
let $q_i(h_j) = \lbrace 0 \rbrace$ for all $j$.  
For worse case choice of $h*$, we need ask every question
$q_i$ for $i \in {1  ... |H|}$ in order to identify $h^*$.  
However, if we define the objective to be
\[ F_{h}(\hat{S}) \triangleq I((q_{|H| + 1}, 0) \in \hat{S}) \]
for all $h$ with $\alpha  = 1$, 
the interactive submodular set cover problem is easy.  To satisfy
$F_{h^*}(\hat{S}) \geq \alpha$ we simply need to ask question
$q_{|H| + 1}$.  By making the cost of $q_{|H| + 1}$ small and
the cost of the other questions large, we get an approximation
ratio of at least $|H| \max_i c(q_i) / \min_i c(q_i)$. 
\end{proof}

\subsection{Adaptivity Gap}

Another simple approach is to ignore feedback and solve the covering
problem for all $h \in H$.  We call this the Cover All method.  This
method is an example of a non-adaptive method: a non-adaptive (i.e.\
non interactive) method is any method that does not use responses to
previous questions in deciding which question to ask next.  The
\emph{adaptivity gap} \cite{approxstochknap} for a problem
characterizes how much worse the best non-adaptive method can perform
as compared to the best adaptive method.  For interactive
submodular set cover we define the adaptivity gap to be the maximum
ratio between the cost of the optimal non-adaptive strategy and the
optimal adaptive strategy.  With this definition, we can show that, in
contrast to related problems \cite{maximizingstochastic} where the
adaptivity gap is a constant, the adaptivity gap for interactive
submodular set cover is quite large.

\begin{thm}
The adaptivity gap for interactive submodular set cover is at least
$\Omega(|H| / \ln |H|)$.
\label{coverallgap}
\end{thm}

\begin{proof}
The result follows directly from the connection to active learning
(Section \ref{connectiontoactivesec}) and in particular any example
of exact active learning giving an exponential speed up over
passive learning.  A classic example is learning a threshold
on a line \cite{analysisofgreedyactive}.  Let $|H| = 2^k$ for
some integer $k > 0$.  Define the active learning objective as before
\[ F_{h}(\hat{S}) \triangleq |H \setminus V(\hat{S})| \] for all $h$.
The goal of the problem is to identify $h^*$.  We define the
query set such that we can identify $h^*$ through binary search.  Let
there be a query $q_i$ corresponding to each hypothesis $h_i$.  Let
$q_i(h_j) = \lbrace 1 \rbrace$ if $i \leq j$ and $q_i(h_j) = \lbrace 0
\rbrace$ if $i > j$.  Each $q_i$ can be thought of as a point on a
line with $h_i$ the binary classifier which classifies all points as
positive which are less than or equal to $q_i$.  By asking question
$q_{2^{k-1}}$ we can eliminate half of $H$ from the version space.  We
can then recurse on the remaining half of $H$ and identify $h^*$ in
$k$ queries.  Any non-adaptive strategy on the other hand must perform
all $2^k$ queries in order to ensure $V(\hat{S})| = 1$ for worst case
choice of $h^*$.
\end{proof}

This result shows, even if we optimally solve the submodular set
cover problem, the Cover All method can incur exponentially greater
cost than the optimal adaptive strategy.

\subsection{Hardness of Approximation}

We show that the $1 + \ln (\alpha |H|)$ approximation factor achieved
by the method we propose is in fact the best possible up to the
constant factor assuming there are no slightly superpolynomial time
algorithms for NP.  The result and proof are very similar to those for
the non-interactive setting \cite{robust}.

\begin{thm}
Interactive submodular set cover cannot be approximated within a
factor of $(1 - \epsilon) \max(\ln |H|, \ln \alpha)$ in polynomial
time for any $\epsilon > 0$ unless NP has $n^{O(\log \log n)}$ time
deterministic algorithms.
\end{thm}

\begin{proof}
We show the result by reducing set cover to interactive submodular
set cover in two different ways.  In the first reduction, a set
cover instance of size $n$ gives an interactive submodular set cover
of with $|H|=1$ and $\alpha=n$. In the second reduction, a set cover instance 
of  size $n$ gives an interactive submodular set cover instance with $|H|=n$
and $\alpha = 1$.  The theorem then follows from the result of
\citet{thresholdsetcover} which shows a set cover 
cannot be approximated within a factor of $(1 - \epsilon) \ln n$ 
in polynomial time for any $\epsilon > 0$ unless NP has $n^{O(\log \log n)}$ time
deterministic algorithms.

Let $V$ be the set of sets defining the set cover problem.  The ground
set is $\bigcup_{v \in V} v$.  The goal of set cover is to find a
small set of sets $S \subseteq V$ such that $\bigcup_{s \in S} s =
\bigcup_{v \in V} v$.  For both reductions we use $|R|=1$ (all
questions have only one response) and make each question in $Q$
correspond to a set in $V$. For a set of question-response pairs
$\hat{S}$ define $V_{\hat{S}}$ to be the subset of $V$ corresponding
to the questions in $\hat{S}$.  For the first reduction with $|H|=1$,
we set the one objective function $F_h(\hat{S})
\triangleq |\bigcup_{v \in V_{\hat{S}}} v|$.  With $\alpha=n$, we have
that $\bar{F}_{\alpha}(\hat{S}) = \alpha$ iff $V_{\hat{S}}$ forms a cover.

For the second reduction with $|H|=n$, define $F_{h_i}(\hat{S})$ for
the $i$th hypothesis $h_i$ to be $1$ iff the $i$th object in the
ground set of the set cover problem is covered by $V_{\hat{S}}$.  More
formally $F_{h_i}(\hat{S}) \triangleq I(v_i \in V_{\hat{S}})$ where
$v_i$ is the $i$th item in the ground set (ordered arbitrarily).  This
is similar to the first reduction except we have broken down the
objective into a sum over the ground set elements. With
$\alpha=1$, we then have that $\bar{F}_{\alpha}(\hat{S}) = \alpha$ iff
$V_{\hat{S}}$ forms a cover.
\end{proof}

The approximation factor we have shown for the greedy algorithm
is
\[ 1 + \ln(\alpha |H|) = 1 + \ln \alpha + \ln |H| < 1 + 2 \max(\ln |H|,
\ln \alpha) \] so our hardness of approximation result
matches up to the constant factor and lower order term.

\section{Experiments}

\begin{table*}[t]
\begin{small}
\begin{center}
\begin{tabular}{|l|r|r|r|}
\hline 
Data Set / Hypothesis Class & Simultaneous Learning and Covering & 
Learn then Cover & Cover All \\
\hline
Enron / Clusters & \textbf{156.64} & 161.81 & 3091.00 \\
Physics / Clusters & \textbf{175.97} & \textbf{177.88} & 3340.00 \\
Physics Theory / Clusters & \textbf{172.38} & \textbf{175.12} & 3170.00 \\
Epinions / Clusters & \textbf{774.81} & 779.23 & 15777.00 \\
Slashdot / Clusters & \textbf{709.30} & 715.39 & 15383.00 \\
\hline 
Enron / Noisy Clusters & \textbf{179.00} & 231.03 & 3091.00 \\
Physics / Noisy Clusters & \textbf{186.13} & 225.02 & 3340.00 \\
Physics Theory / Noisy Clusters & \textbf{160.62} & 201.24 & 3170.00 \\
Epinions / Noisy Clusters & \textbf{788.52} & \textbf{788.06} & 15777.00 \\
Slashdot / Noisy Clusters & \textbf{804.87} & \textbf{804.86} & 15383.00 \\
\hline
\end{tabular}
\end{center}
\end{small}
\caption{Average number of queries required to find a dominating set in
the target group.}
\label{results}
\end{table*}

We tested our method on the interactive dominating set problem
described in Section \ref{examplesec}.  In this problem, we are given
a graph and $H$ is a set of possibly overlapping clusters of nodes.
The goal is to find a small set of nodes which forms a dominating set
of an initially unknown target group $h^* \in H$.  After selecting
each node, we receive feedback indicating if the selected node is in
the target group.  Our proposed method (Simultaneous Learning and Covering)
simultaneously learns about the target group $h^*$ and finds a
dominating set for it.  We compare to two baselines: a method which
first exactly identifies $h^*$ and then finds a dominating set for the
target group (Learn then Cover) and a method which simply ignores
feedback and finds a dominating set for the union of all clusters
(Cover All).  Note that Theorem \ref{learnthencovergap} and Theorem
\ref{coverallgap} apply to Learn then Cover and Cover All
respectively, so these methods do not have strong theoretical
guarantees.  However, we might hope however that for reasonable real
world problems they perform well.  We use real world network data sets
with simple synthetic hypothesis classes designed to illustrate
differences between the methods.  The networks are from Jure
Leskovec's collection of datasets available at
\url{http://snap.stanford.edu/data/index.html}. We convert all the
graphs into undirected graphs and remove self edges.
 
Table \ref{results} shows our results.  Each reported result is the
average number of queries over 100 trials.  Bolded results are the
best methods for each setting with multiple results bolded when
differences are not statistically significant (within $p=.01$ with a
paired t-test).  In the first set of results (Clusters), we create $H$
by using the METIS graph partition package 4 separate times
partitioning the graph into 10, 20, 30, and 40 clusters.  $H$ is the
combined set of 100 clusters, and these clusters overlap since they
are taken from 4 separate partitions of the graph.  The target $h^*$
is chosen at random from $H$.  With this hypothesis class, we've found
that there is very little difference between the Simultaneous Learning and
Covering and the Learn then Cover methods.  The Cover All method performs
significantly worse because without the benefit of feedback it must
find a dominating set of the entire graph.

In the second set of results, we use a hypothesis class designed to
make learning difficult (Noisy Clusters).  We start with $H$ generated as
before.  We then add to $H$ 100 additional hypotheses which are each
very similar to $h^*$.  Each of these hypotheses consists of the
target group $h^*$ with a random member removed.  $H$ is then the
combined set of the 100 original hypotheses and these 100 variations
of $h^*$.  For this hypothesis class, Learn then Cover performs
significantly worse than our Simultaneous Learning and Covering method
on 3 of the 5 data sets.  Learn then Cover exactly identifies $h^*$,
which is difficult because of the many hypotheses similar to $h^*$.
Our method learns about $h^*$ but only to the extent that it is
helpful for finding a small dominating set.  On the other two data
sets Learn then Cover and Simultaneous Learning and Covering are
almost identical.  These are larger data sets, and we've found that
when the covering problem requires many more queries than the learning
problem, our method is nearly identical to Learn then Cover.
This makes sense since when $\alpha$ is large compared to the sum over
$F_h(\hat{S})$ the second term in $\bar{F}_{\alpha}$ dominates.

It is also possible to design hypothesis classes for which Cover All
outperforms Learn then Cover: we found this is the case when the
learning problem is difficult but the subgraph corresponding to the
union of all clusters in $H$ is small.  In the appendix we give an
example of this.  In all cases, however, our approach does about as
good or better than the best of these two baseline methods.
Although we use real world graph data, the hypothesis classes and
target hypotheses we use are very simple and synthetic, and as such these
experiments are primarily meant to provide reasonable examples in support
of our theoretical results.

\section{Future Work}

We believe there are other interesting applications which can be posed
as interactive submodular set cover.  In some applications it may be
difficult to compute $\bar{F}_{\alpha}$ exactly because $H$ may be
very large or even infinite.  In these cases, it may be possible to
approximate this function by sampling from $H$.  It's also important
to consider methods that can handle misspecified hypothesis classes
and noise within the learning. One approach could be to extend
agnostic active learning \cite{agnostic} results to a similar
interactive optimization setting.

\begin{small}
\bibliography{maxunknowneetr}

\begin{thebibliography}{18}
\providecommand{\natexlab}[1]{#1}
\providecommand{\url}[1]{\texttt{#1}}
\expandafter\ifx\csname urlstyle\endcsname\relax
  \providecommand{\doi}[1]{doi: #1}\else
  \providecommand{\doi}{doi: \begingroup \urlstyle{rm}\Url}\fi

\bibitem[Asadpour et~al.(2008)Asadpour, Nazerzadeh, and
  Saberi]{maximizingstochastic}
A.~Asadpour, H.~Nazerzadeh, and A.~Saberi.
\newblock Stochastic submodular maximization.
\newblock In \emph{Workshop on Internet and Network Economics}, 2008.

\bibitem[Balcan et~al.(2006)Balcan, Beygelzimer, and Langford]{agnostic}
M.~Balcan, A.~Beygelzimer, and J.~Langford.
\newblock {Agnostic active learning}.
\newblock In \emph{ICML}, 2006.

\bibitem[Balc\'{a}zar et~al.(2007)Balc\'{a}zar, Castro, Guijarro, K\"{o}bler,
  and Lindner]{generaldim}
J.~Balc\'{a}zar, J.~Castro, D.~Guijarro, J.~K\"{o}bler, and W.~Lindner.
\newblock A general dimension for query learning.
\newblock \emph{Journal of Computer and System Sciences}, 73\penalty0
  (6):\penalty0 924--940, 2007.

\bibitem[Dasgupta(2004)]{analysisofgreedyactive}
S.~Dasgupta.
\newblock Analysis of a greedy active learning strategy.
\newblock In \emph{NIPS}, 2004.

\bibitem[Dasgupta et~al.(2003)Dasgupta, Lee, and Long]{querycollab}
S.~Dasgupta, W.~Lee, and P.~Long.
\newblock A theoretical analysis of query selection for collaborative
  filtering.
\newblock \emph{Machine Learning}, 51\penalty0 (3), 2003.

\bibitem[Dean et~al.(2004)Dean, Goemans, and Vondrak]{approxstochknap}
B.~Dean, M.~Goemans, and J.~Vondrak.
\newblock {Approximating the stochastic knapsack problem: The benefit of
  adaptivity}.
\newblock In \emph{FOCS}, 2004.

\bibitem[Feige(1998)]{thresholdsetcover}
U.~Feige.
\newblock A threshold of ln n for approximating set cover.
\newblock \emph{Journal of the ACM}, 45\penalty0 (4), 1998.

\bibitem[Goemans and Vondr{\'a}k(2006)]{stochasticcovering}
M.~Goemans and J.~Vondr{\'a}k.
\newblock Stochastic covering and adaptivity.
\newblock In \emph{LATIN}, 2006.

\bibitem[Golovin and Krause(2010)]{adaptsubmod}
D.~Golovin and A.~Krause.
\newblock Adaptive submodularity: A new approach to active learning and
  stochastic optimization.
\newblock In \emph{COLT}, 2010.

\bibitem[Hanneke(2006)]{costcomp}
S.~Hanneke.
\newblock The cost complexity of interactive learning, 2006.
\newblock Unpublished.
  \url{http://www.stat.cmu.edu/~shanneke/docs/2006/cost-complexity-working-not%
es.pdf}.

\bibitem[Karypis and Kumar(1999)]{metis}
G.~Karypis and V.~Kumar.
\newblock A fast and highly quality multilevel scheme for partitioning
  irregular graphs.
\newblock \emph{SIAM Journal on Scientific Computing}, 1999.

\bibitem[Kempe et~al.(2003)Kempe, Kleinberg, and Tardos]{maximizingspread}
D.~Kempe, J.~Kleinberg, and E.~Tardos.
\newblock Maximizing the spread of influence through a social network.
\newblock In \emph{KDD}, 2003.

\bibitem[Kempe et~al.(2005)Kempe, Kleinberg, and Tardos]{influentialnodes}
D.~Kempe, J.~Kleinberg, and E.~Tardos.
\newblock Influential nodes in a diffusion model for social networks.
\newblock In \emph{ICALP}, 2005.

\bibitem[Krause et~al.(2008)Krause, McMahan, Guestrin, and Gupta]{robust}
A.~Krause, H.~McMahan, C.~Guestrin, and A.~Gupta.
\newblock Robust submodular observation selection.
\newblock \emph{JMLR}, 2008.

\bibitem[Lin and Bilmes(2010)]{documentsum}
H.~Lin and J.~Bilmes.
\newblock Multi-document summarization via budgeted maximization of submodular
  functions.
\newblock In \emph{NAACL/HLT}, 2010.

\bibitem[Narayanan(1997)]{electrical}
H.~Narayanan.
\newblock \emph{Submodular Functions and Electrical Networks}.
\newblock North Holland, 1997.

\bibitem[Streeter and Golovin(2008)]{onlinemaximizing}
M.~Streeter and D.~Golovin.
\newblock An online algorithm for maximizing submodular functions.
\newblock In \emph{NIPS}, 2008.

\bibitem[Wolsey(1982)]{analysisofgreedy}
L.~Wolsey.
\newblock {An analysis of the greedy algorithm for the submodular set covering
  problem}.
\newblock \emph{Combinatorica}, 2\penalty0 (4), 1982.

\end{thebibliography}
\end{small}
\bibliographystyle{abbrvnat}

\appendix

\section{Additional Experiments}

\label{appendix}

Table \ref{moreresults} shows additional experimental results using
different hypothesis classes. In the first set of results, we use a
hypothesis class $H$ consisting of 100 randomly chosen geodesic balls
of radius 2 (Balls).  Each group $h \in H$ is formed by choosing a
node uniformly at random from the graph and then finding all nodes
within a shortest path distance of 2.  The target group $h^*$ is then
selected at random from $H$.  With this hypothesis class, we've found
that there is very little difference between the Simultaneous Learning
and Covering and the Learn then Cover methods, similar to the Clusters
hypothesis class in Table \ref{results}.  Learn then Cover is better
on 3 of the 5 data sets, but the difference is very small (around 1
query).  The Cover All method again performs significantly worse
because it must find a dominating set of all 100 of the geodesic
balls.

In the second set of results, Noisy Balls, we use a hypothesis class
similar to the Noisy Clusters hypothesis class in Table \ref{results}
but using random geodesic balls.  We first generate 2 core groups by
sampling random geodesic balls of radius 2 as before.  We then
generate 50 small variations of each of these 2 core groups, each
consisting of the core group with a random member removed.  $H$ is
this set of 100 variations, and the target group $h^*$ is again
selected at random from $H$.  For this hypothesis class, Simultaneous
Learning and Covering outperforms the other methods because it learns
about $h^*$ but only to the extent that it is helpful for finding a
small dominating set.  Cover All actually outperforms Learn then Cover
with this hypothesis class, because the total number of vertices in
the union of all clusters in $H$ is small.

In the third set of results denoted Expanded Clusters, we create $H$
by partitioning the graph into 100 clusters using the METIS
\cite{metis} graph partitioning package and then expand each of these
clusters to include its immediate neighbors.  This creates a set of
100 overlapping clusters with shared vertices on the fringes of each
cluster.  As before the target hypothesis is selected at random from
$H$.  We have found that results with this hypothesis class are similar
to those with the Balls and Clusters hypothesis class.

\begin{table*}[t]
\begin{small}
\begin{center}
\begin{tabular}{|l|r|r|r|}
\hline 
Data Set / Hypothesis Class & Simultaneous Learning and Covering & 
Learn then Cover & Cover All \\
\hline
Enron / Balls & 15.37 & \textbf{14.29} & 390.60 \\
Physics / Balls & \textbf{28.83} & \textbf{28.84} & 1096.58 \\
Physics Theory / Balls & \textbf{28.74} & \textbf{28.96} & 826.55 \\
Epinions / Balls  & 19.53 & \textbf{18.37} & 829.69 \\
Slashdot / Ball & 18.32 & \textbf{17.73} & 952.09 \\
\hline 
Enron / Noisy Balls & \textbf{8.36} & 27.11 & 14.25 \\
Physics / Noisy Balls & \textbf{21.47} & 41.03 & 38.31 \\
Physics Theory / Noisy Balls & \textbf{22.48} & 44.82 & 37.43 \\
Epinions / Noisy Balls & \textbf{15.03} & 32.76 & \textbf{18.11} \\
Slashdot / Noisy Balls & \textbf{12.81} & 32.53 & 31.38 \\
\hline
Enron / Expanded Clusters & \textbf{84.90} & \textbf{84.23} & 3091.00 \\
Physics / Expanded Clusters & \textbf{150.28} & 152.21 & 3340.00 \\
Physics Theory / Expanded Clusters & \textbf{120.84} & 122.12 & 3170.00 \\
Epinions / Expanded Clusters & \textbf{260.21} & 261.01 & 15777.00 \\
Slashdot / Expanded Clusters & \textbf{324.15} & 325.35 & 15383.00 \\
\hline 
\end{tabular}
\end{center}
\end{small}
\caption{Average number of queries required to find a dominating set in the target community.}
\label{moreresults}
\end{table*}

\end{document}